\documentclass[twoside,leqno,twocolumn]{article}  
\usepackage{amsmath}
\usepackage{graphicx}\usepackage{ifpdf}
\usepackage{subfigure}\usepackage{algorithm2e}

\newcommand{\remove}[1]{} % comment out text

\begin{document}

\title{\Large Transfer Topic Modeling with Ease and Scalability}
\author{Jeon-Hyung Kang, Jun Ma, Yan Liu \\
Computer Science Department\\
University of Southern California\\
Los Angeles, CA 90089\\
\texttt{jeonhyuk,junma,yanliu@usc.edu}}
\date{}

\maketitle

%\pagenumbering{arabic}
%\setcounter{page}{1}%Leave this line commented out.

\begin{abstract} 
The increasing volume of \textit{short} texts generated on social media sites, such as Twitter or Facebook, creates a great demand for effective and efficient topic modeling approaches.  While latent Dirichlet allocation (LDA) can be applied, it is not optimal due to its weakness in handling short texts with fast-changing topics and scalability concerns. 
In this paper, we propose a transfer learning approach that utilizes abundant labeled documents from other domains (such as Yahoo! News or Wikipedia) to improve topic modeling, with better model fitting and result interpretation.  Specifically, we develop \emph{Transfer Hierarchical} LDA (thLDA) model, which incorporates the label information from other domains via informative priors. In addition, we develop a parallel implementation of our model for large-scale applications.  We demonstrate the effectiveness of our thLDA model on both a microblogging dataset and standard text collections including AP and RCV1 datasets.
\end{abstract}

\section{Introduction}
Social-media websites, such as Twitter and Facebook, have become a novel real-time channel for people to share information on a broad range of subjects. 
With millions of messages and updates posted daily, there is an obvious need for effective ways to organize the data tsunami.
Latent Dirichlet Allocation (LDA), as a Bayesian hierarchical model to capture the text generation process \cite{blei2003latent}, has been shown to be very powerful in modeling text corpus. However, several major characteristics distinguish social media data from traditional text corpus and raise great challenges to the LDA model.  First, each post or tweet is
limited to a certain number of characters, and as a result abbreviated syntax is often introduced.  Second, the texts are very noisy, with broad topics and repetitive, less meaningful content. Third, the input data typically arrives in high-volume streams. 

It is known that the topic outputs by LDA completely depend on the word distributions in the training documents. Therefore, LDA results on blog and microblogging data would naturally be poor, with a cluster of words that co-occur in many documents without actual semantic meanings \cite{puniyani2010social, wallach2009rethinking}. 
Intuitively, the generative process of LDA model can be guided by the document labels so that the learned hidden topics can be more meaningful. Therefore, in \cite{blei2008supervised, lacoste2008disclda, ramage2009labeled}, discriminative training of LDA has been explored and in particular, \cite{RamageD2010Characterizing} applied \emph{labeled} LDA (lLDA) to analyze Twitter data by using hashtags as labels. This approach addresses our challenges to a certain extent. Even though supervised LDA gives more comprehensible topics than LDA, it has the general limitation of the LDA framework in that each document is represented by a single topic distribution. This makes comparing documents difficult when we have sparse, noisy documents with constantly changing topics. Furthermore, it is very difficult to obtain the labels (e.g.  hashtags) for continuously growing text collections like social media, not to mention the fact that on Twitter application, many hashtags refer to very broad topics (e.g. ``\#internet'', ``\#sales'')  and therefore could even be misleading when used to guide the topic models. \cite{blei2004hierarchical, blei2010nested} proposed hierarchical LDA model (hLDA), that generates topic hierarchies from an infinite number of topics, which are represented by a topic tree, and each document is assigned a path from tree root to one of the tree leaves. hLDA has the capability to encode the semantic topic hierarchies,  but when fed with noisy and sparse data such as the user generated short tweet messages, it is not robust and its results lack meaningful interpretations \cite{puniyani2010social, wallach2009rethinking}. 

Recently, hLDA has been studied in  \cite{Celikyilmaz} to summarize multiple documents. They built a two-level learning model using hLDA model to discover similar sentences to the given summary sentences using hidden topic
distributions of the document. The distance dependent CRP \cite{Blei_distancedependent} studied several types of decay: window decay,
exponential decay, and logistic decay, where customers' table assignments depend on external distances between them. Our paper aims to bridge the gap between short, noisy texts and their actual generation process without additional labeling efforts. At the same time, we develop parallel algorithms to speed up inference  so that our model can be applicable to large-scale applications.

In this paper, we propose a simple solution model, transfer hierarhicial LDA (thLDA) model. The basic idea is extracting human knowledge on topics  from source domain corpus in the form of representative words that are consistently meaning across contexts or media and encode them as priors to learn the topic hierarchies in target domain corpus of the hLDA model. To extract source domain corpus, thLDA model utilizes related labeled documents from other sources (such as Yahoo! news or socially tagged web pages) and to encode the laels, we modified a nested Chinese Restaurant Process (nCRP) as guidance for inferring latent topics of target domains. We base our model on hierarchical LDA (hLDA) \cite{blei2004hierarchical, blei2010nested} mainly because hLDA has the natural capability to encode the semantic topic hierarchies with document clusters. In addition, recent study suggests that hierarchical Dirichlet process provides an effective explanation model for human transfer learning \cite{Canini_modelingtransfer}.

The rest of the paper is organized as follows:  In Section 2, we describe the proposed
methodology in detail and discuss its relationship with existing
work.  In Section 3, we demonstrate the effectiveness of our model,
and summarize the work and discuss future directions in Section 4.

\section{Related Work}

\subsection{Topic Models}

Latent Dirichlet Allocation (LDA) \cite{blei2003latent} has gained popularity for automatically extracting a representation of corpus. LDA is a completely unsupervised model that views documents as a mixture of probabilistic topics that is represented as a K dimensional random variable $\theta$. 
In generative story, each document is generated by first picking a topic distribution $\theta$ from the Dirichlet prior and then use each document's topic distribution $\theta$ to sample latent topic variables $z_i$. LDA makes the assumption that each word is generated from one topic where $z_i$ is a latent variable indicating the hidden topic assignment for word $w_i$. The probability of choosing a word $w_i$ under topic $z_i$, $p(w_i|z_i,\beta)$, depends on different documents. LDA is not appropriate for labeled corpora, so it has been extended in several ways to incorporate a supervised label set into its learning process. In \cite{ramage2009labeled}, Ramage et al. introduced \emph{Labeled} LDA, a novel model that use multi-labeled corpora to address the credit assignment problem. Unlike LDA, \emph{Labeled} LDA constrains topics of documents to a given label set. 
Instead of using symmetric Dirichlet distribution with a single hyper-parameter $\alpha$ as a Dirichlet prior on the topic distribution $\theta_{(d)}$, it restricted $\theta_{(d)}$ to only the topics that correspond to observed labels $\Lambda_{(d)}$. 
\remove{
\begin{figure}[tb]
\begin{center}
\begin{tabular}{@{}c@{}}
\includegraphics[width=0.6\linewidth]{LDA.png}\\
(a) \\ %Type 1\\
\includegraphics[width=0.6\linewidth]{LLDA.png}\\
(b) \\%Type 2
\includegraphics[width=0.6\linewidth]{hlda.png}\\
(c) %hlda
\end{tabular}
\end{center}
\caption{Graphical models: (a) LDA (b) \emph{Labeled} LDA. (c) \emph{Hierarchical} LDA. }
\label{fig:ldallda}
\end{figure}
}%\paragraph{Hierarchical LDA}
In \cite{blei2004hierarchical,blei2010nested}, the authors proposed a stochastic processes, where the Bayesian inference are no longer restricted to finite dimensional spaces. Unlike LDA, it does not restrict the given number of topics and allows arbitrary breadth and depth of topic hierarchies. The topics in hLDA model are represented by a topic tree, and each document is assigned a path from tree root to one of the tree leaves. Each document is generated by first sampling a path along the topic tree, and then sampling topic $z_i$ among all the topic nodes in the path for each word $w_i$. The authors proposed a nested Chinese restaurant process prior on path selection.

\begin{eqnarray}
p(occupied\ table\ i\ |\ previous\ customers) \nonumber\\
 = \frac{n_i}{\gamma + n - 1} \\
p(unoccupied\ table\ |\ previous\ customers)\nonumber\\
 = \frac{\gamma}{\gamma + n - 1}
\label{equ:crpb}
\end{eqnarray}

The nested Chinese restaurant process implies that the first customer sits at the first table and the nth customer sits at a table i which is drawn from the above equations. When a path of depth d is sampled, there are d number of topic nodes along the path, and the document sample the topic $z_i$ among all topic nodes in the path for each word $w_i$ based on GEM distribution. The experiment result of hLDA shows that the above document generating story can actually encode the semantic topic hierarchies.

\subsection{Transfer Learning} 
Transfer learning has been extensively studied in the past decade to
leverage the data (either labeled or unlabeled) from one task (or
domain) to help another  \cite{pan2009survey}. In summary, there
are two types of transfer learning problems, \textit{i.e.} shared
label space or shared feature space. For shared label space
\cite{arnold2008comparative}, the main objective is to transfer the
label information between observations from different distributions
(i.e. domain adaptation) or uncover the relations between multiple
labels for better prediction (i.e. multi-task learning); for shared
feature space, one of the most representative works is self-taught
learning \cite{raina2007self}, which uses sparse coding
\cite{lee2007efficient} to construct higher-level features via
abundant unlabeled data to help improve the performance of the
classification task with a limited number of labeled examples.

As an unsupervised generative model, LDA possesses the advantage of modeling the generative procedure of the whole dataset by
establishing the relationships between the documents and their associated hidden topics, and between the hidden topics and the
concrete words,  in an un-supervised way. Intuitively, transfer learning on this generative model can be realized in two ways: one is utilizing the document labels from the other domain (with the assumption that the target domain and source domain share the same label space) so that the learned hidden topics can be much more meaningful, and the other is utilizing the documents from other domains to enrich the contents so that we can learn a more robust shared latent space. %

 In \cite{blei2008supervised, lacoste2008disclda, ramage2009labeled},
the authors proposed the discriminative LDA, which adds supervised information to the original LDA model, and guides the
generative process of documents by using the labels. 
These methods clearly demonstrate the advantages of the discriminative training of generative models. However, this is different from transfer learning since they simply utilize the labeled documents in the same domain to help build the generative model.  But the same motivation can be applied to transfer learning, in which the supervised information is used to guide the generation of the common latent semantic space shared by both the source domain and the target domain. Transferring information from source to target domain is extremely desirable in social media analysis, in which the target domain example features are very sparse, with a lot of missing features. Based on the shared common latent semantic space, the missing features can be recovered to some extent, which is helpful in better representing these examples.

\section{Transfer Topic Models}

Content analysis on social media data is a challenging problem due to its unique language characteristics, which prevents standard text mining tools from being used to their full potential. Several models have been developed to overcome this barrier by aggregating many messages~\cite{Hong10empiricalstudy},  applying temporal attributes~\cite{michelson:twitter, Ritter10unsupervisedmodeling, Tian:2010:TDO:1871437.1871752}, examining entities~\cite{michelson:twitter} or applying manual annotation to guide the topic generation~\cite{RamageD2010Characterizing}. 
The main motivation of our work is that previous unsupervised approaches for analyzing social media data fail to achieve high accuracy due to the noise and sparseness, while supervised approaches require annotated data, which often requires a significant amount of human effort. Even though \emph{hierarchical} LDA has the natural capability to encode the semantic topic hierarchies with clusters of similar documents based on that hierarchies, it still cannot provide robust result for noise and sparseness (Fig \ref{fig:hlda-tree}) because of the exchangeability assumption in Dirichlet Process mixture \cite{Blei_distancedependent} . Exchangeability is essential for CRP mixture to be equivalent to a DP mixture; thus customers are exchangeable. However this assumption is not reasonable when it is applied to microblogging dataset. Based on our experiment, when the data set is noisy and sparse, unrelated words tends to cluster with other word because of co-occurrences.(Fig \ref{fig:hlda-tree}) 

\subsection{Knowledge Extraction from Source Domain}

Consider the task of modeling short and sparse documents based on specific source domain structure. For example, a user who is interested in browsing documents with particular categories of topics might prefer to see clusters of other documents based on his category. Clustering target domain by transferring his topic hierarchy category in the source domain, we could produce better document clusters and topic hierarchies by leveraging the context information from the source domain. 

User generated categories can be found from various source domains: Twitter list, Flickr collection and set, \textit{Del.icio.us} %delicious.com
hierarchy and Wikipedia or News categories. We transfer source domain knowledge to target domain documents by assigning a prior path, sequence of assigned topic nodes from root to leaf node. The prior paths of the documents can be used to identify whether two target documents are similar or not, so that our model could group similar documents cluster together while keep different documents separate based on the label.  To label each document's prior path on the source domain hierarchy, we generate word vectors of nodes in the source domain hierarchy. Each label is generated by measuring the similarity between source domain topic hierarchy and document. There are many ways to measure similarity between two vectors: cosine similarity, euclidean distance, Jaccard index and so on.  For simplicity, in this paper, we label our prior knowledge of the target document by computing cosine similarity between the target document and a node in the source domain hierarchy. We start from the root node of the hierarchy and keep assigning only the most similar topic node at each level while only considering child nodes of currently assigned topic nodes as next level candidates.

\begin{figure}[tb]
\begin{center}
\begin{tabular}{cc}
\includegraphics[width=0.7\linewidth]{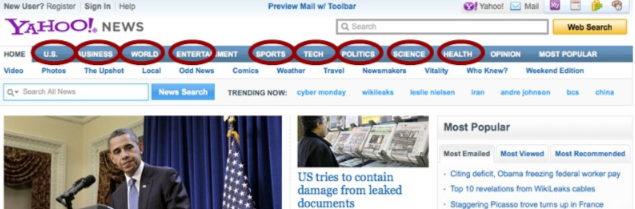}\\
(a)\\
\includegraphics[width=0.75\linewidth]{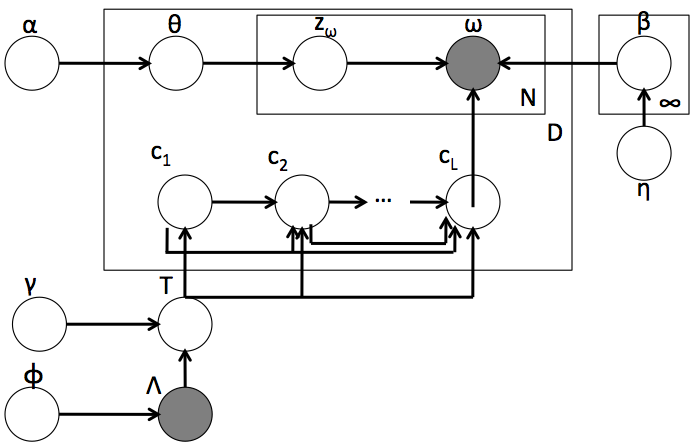}\\
(b)\\
\end{tabular}
\end{center}
\caption{(a)Yahoo! news home page and the categories of news (b) \emph{The graphical model representation of transfer Hierarchical LDA with a modified-nested CRP prior} }
\label{fig:lhlda}
\end{figure}
\subsection{Transfer Hierarchical LDA}
We incorporated the label hierarchies into the model by changing the prior of the path in hLDA in a way that the path selection favors the ones in the existing label hierarchies. Similar to original hLDA, thLDA models each document, a mixture of topics on the path, and generates each word from one of the topics. In the original hLDA model, the path prior is nested \emph{Chinese Restaurant Process} (nCRP), where the probability of choosing one topic in one topic layer depends on the number of documents already assigned to each node in that layer. That is, nodes assigned with more documents will have a higher probability of generating new documents. However, thLDA incorporates supervision by modifying path prior using following equation:
\begin{eqnarray}
p(table\ i\ without\ prior\ |\ previous\ customers) \nonumber\\
 = \frac{n_i}{\gamma + k\lambda+ n} \\
p(table\ i\ with\ \ prior\ |\ previous\ customers)\nonumber\\
 = \frac{n_i + \lambda}{\gamma + k\lambda + n} \\
p(unoccupied\ table\ |\ previous\ customers)\nonumber\\
 = \frac{\gamma}{\gamma + k\lambda+ n}
\label{equ:lcrp3}
\end{eqnarray}
where k is the total number of labels for the document, d is the total number of documents, and $\lambda$ is the weighted indicator variable that controls the strength of a prior added into the original nested Chinese restaurant process. The graphical model of the new model is shown in Fig \ref{fig:lhlda} (b). Whether the customer sits at a specific table is not only related to how many customers are currently sitting at table i (${n_i}$) and how often a new customer chooses a new table  ($\gamma$), but is also related to how close the current customer is to customers at table i ($\lambda$). Note that, in this work we use same $\lambda$ for different topic for simplicity; however table specific $\lambda$ or different prior for different table can be applied for a sophisticated model.

In the transfer hierarchical LDA model, the documents in a corpus are assumed to be drawn from the following generative process:
\begin{tabbing}
(1) For \=each table k in the infinite tree\\
\; \;  \;   (a) Generate \= $\beta_{(k)} \sim Dirichlet({\eta}$) \\
(2) For \=each document d \\
\; \;   \;  (a) Generate \= $c_{(d)} \sim$ modified nCRP $({\gamma, \lambda}$) \\
\; \;   \;  (b) Generate  \=  $\theta_{(d)} | {\{m,\pi\}} \sim GEM ({m, \pi})$\\ 
\; \;  \;  (c) For each word, \\  
\; \; \; \;\; \;\; \;  i. Choose level $Z_{d,n} | {\theta_d}  \sim Discrete ({\theta_d})$\\
\; \; \; \;\; \; \; \;ii. Choose word $W_{d,n} | \{{z_{d,n},c_d,\beta}\} $  \\
\>\; \; \; \;\; \; \; \;\; \;$ \sim  Discrete ({\beta_{c_d}[z_{d,n}]})$
\end{tabbing}

The variables notation are: $z_{d,n}$, the topic assignments of the $n$th word in the $d$th document over L topics; $w_{d,n}$, the $n$th word in the $d$th document. In Fig \ref{fig:lhlda} (b), T represents a collection of an infinite number of L level paths which is drawn from modified nested CRP. $c_{m,l}$ represents the restaurant correponding to the $l$th topic distribution in $m$th document and distribution of ${c_{m,l}}$ will be defined by the modified nested CRP conditioned on all the previous $c_{n,l}$ where $n<m$. We assume that each table in a restaurant  is assigned a parameter vector $\beta$ that is a multinomial topic distribution over vocabulary for each topic $k$ from a Dirichlet prior $\eta$. We also assume that the words are generated from a mixture model which is a  specific random distribution of each document. A document is drawn by first choosing an L level path through the modified-nested CRP and then drawing the words from the L topics which are associated with the restaurants along the path. $\Lambda^{(d)}$ = ($l_1$, ..., $l_k$) refers binary label presence indicators and $\Phi_{k}$ is the label prior for label k. 

%\textsc{Can we say something similar as eq(2.1-2.3) for transfer lda?}
thLDA is able to transfer source domain knowledge to topic hierarchy by making an assignment related to not only how many documents are assigned in topic  i (${n_i}$) but also how close current topic is to the documents in topic i based on the source domain knowledge ($\lambda$).  For unseen topics from source domain, we do not have knowledge to transfer from source domain, so we assign probability proportional to the number of documents already assigned in topic i. Unlike transferring knowledge only for labeled entities or given source domain knowledge, our model learns both unlabeled and labeled data based on different prior probability equations (4), (5) and (6).  

A modified nested Chinese restaurant process can be imagined through the following scenario. As in \cite{blei2004hierarchical,blei2010nested}, suppose that there are an infinite number of infinite-table Chinese restaurants in a city and there is only one headquarter restaurant. There is a note on each table in every restaurant which refers to  other restaurants and each restaurant is referred to once. So, starting from the headquarter restaurant, all other restaurants are connected as a branching tree. One can think of the table as a door to other restaurants unless the current restaurant is the leaf node restaurant of the tree. So, starting from the root restaurant, one can reach the final destination table of leaf node restaurant, and the customers in the same restaurant share the same path. When a new customer arrives, instead of following the original nested chinese restaurant process, we put a higher weight on the table where similar customers are seated. 

\subsection{Gibbs Sampling for Inference}

The inference procedure in our thLDA model is similar to hLDA except for the modification of the path prior. We use the gibbs sampling scheme: for each document in the corpus, we first sample a path $c_d$ in the topic tree based on the path sampling of the rest of the documents $c_{-d}$ in the corpus:
\begin{equation}
\begin{aligned}
p(\bf{c_d|w,c_{-d},z},\eta,\gamma,\lambda) \\
\propto 
 p(\bf{c_d}|\bf{c_{-d}},\gamma,\lambda)p(\bf{w_d}|\bf{c,w_{-d},z},\eta)
 \end{aligned}
\end{equation}
Here the $p(\bf{c_d}|\bf{c_{-d}},\gamma,\lambda)$ is the prior probability on paths implied by the modified nested CRP, and $p(\bf{w_d}|\bf{c,w_{-d},z},\eta)$ is the probability of the data given for a particular path.

The second step in the Gibbs sampling inference for each document is to sample the level assignments for each word in the document given the sampled path:
\begin{equation}
\begin{aligned}
p(z_{d,n}|\bf{z_{-(d,n)},c,w,}m,\pi,\eta)\\
 \propto 
p(z_{d,n}|\bf{z_{d,-n}},m,\pi)p(w_{d,n}|\bf{z,c,w_{-(d,n)}},\eta)
\end{aligned}
\end{equation}
The first term is the distribution over levels from GEM, and the second term is the word emission probability given a topic assignment for each word. The only thing we need to change in the inference scheme is the path prior probability in the path sampling step.

\subsection{Parallel Inference Algorithm} 

We developed thLDA parallel approximate inference algorithm on independent P processors to facilitate learning efficiency. In other words, we split the data into P parts, and implement thLDA on each processor performing Gibbs sampling on partial data. However, the gibbs sampler requires that each sample step is conditioned on the rest of the sampling states, hence we introduce a tree merge stage to help P Gibbs samplers share the sampling states periodically during the P independent inference processes.

First, given the current global state of the CRP, we sample the topic assignment for word n in document d from processor p:
\begin{equation}
\begin{aligned}
p(z_{d,n,p}|\bf{z_{-(d,n,p)},c,w,}m,\pi,\eta)  \propto \\
p(z_{d,n,p}|\bf{z_{d,-n,p}},m,\pi)p(w_{d,n,p}|\bf{z,c,w_{-(d,n,p)}},\eta)
\end{aligned}
\end{equation}
Here $\bf{z_{-(d,n,p)}}$ is the vectors of topic allocations on process p excluding $\bf{z_{(d,n,p)}}$ and $\bf{w_{-(d,n,p)}}$ is the nth word in document d on process p excluding $\bf{w_{(d,n,p)}}$. Note that on a separate processor, we need to use total vocabulary size and the number of words that have been assigned to the topic in the global state of the CRP.  We merge the P topic assignment count table to a single set of counts after each LDA Gibbs iteration so that the global sampling state is shared among P processes. \cite{newman2007distributed}.
 
Second, on P separate processors, we sample path selection, conditional distribution for ${c_{d,p}}$ given w and c variables for all documents other than d. 
\begin{equation}
\begin{aligned}
p(\bf{c_{d,p}|w,c_{-d,p},z},\eta,\gamma,\lambda) \propto \\
 p(\bf{c_{d,p}}|\bf{c_{-d,p}},\gamma,\lambda)p(\bf{w_{d,p}}|\bf{c,w_{-d,p},z},\eta)
 \end{aligned}
\end{equation}
The conditional distribution for both prior and the likelihood of the data given a particular choice of ${c_{d,p}}$ are computed locally. Note that, to compute the second term it needs to be known the global state of the CRP: documents' path assignment and the number of instances of a word that have been assigned to the topic index by ${c_d}$ on the tree. 
  
To merge topic trees for the global state of the CRP, we first define the similarity between two topics as the cosine similarity of two topics' word distribution vector:
\begin{equation}
similarity_{\bf{\beta_{i}}\bf{\beta_{j}}} = {\bf{{\beta_{i}}}\cdot \bf{{\beta_{j}}}}/ {||\bf{{\beta_{i}}}||  \  ||\bf{{\beta_{j}}}||} 
\end{equation}
For given P number of infinite trees of Chinese restaurants, we pick one tree as the base tree, and recursively merge topics in the remaining P-1 trees into the base tree in a top-down manner. For each topic node $t_i$ being merged, we find the most similar node $t_j$ in the base tree where $t_i$ and $t_j$ have same parent node. 

\begin{algorithm}[]
%\KwIn{sane people and photos}
%\KwOut{bebo addicts}
\ForEach{iteration}{
\
\emph{parallel} \ForEach{document d}{
 \ForEach{word n}{
 sample the topic assignment 
$ p(z_{d,n,p}|\bf{z_{-(d,n,p)},c,w,}m,\pi,\eta)$\\
}}
\ForEach{process p}{
merge topic assignment count table to a single set of counts
}

\emph{parallel}  \ForEach{document d}{
sample path
$p(\bf{c_{d,p}|w,c_{-d,p},z},\eta,\gamma,\lambda) $
using global state of the CRP 
}
Pick one tree q as a base tree in every merge stage, \\
\ForEach{tree from process p $\in$ P-\{q\} }{
\ForEach{depth}{
\ForEach{topic node i}{
find most similar topic node j $\in$ q
\\
\emph{if} parent(i)==parent(j) \\
\ \ merge topics i to topic j \\
\emph{else} \\
\ \ add topic node i to tree q
\
\\
}}
}
}

\caption{The parallel inference algorithm} 

\end{algorithm}

\subsection{Discussion} 

Our thLDA model is different from existing topic models.
In lLDA, they incorporate supervision by restricting $\theta_{d}$ to each document's label set. So word-topic  $Z_{d,n}$ assignments are restricted by its given labels. The number of unique labels K in lLDA is the same as the number of topics K in LDA. Unlike lLDA,  thLDA does not directly correlate label and topic by modifying ${\theta_d}$, so the number of topics are not determined or set by the number of labels, that number serves as a guidance for inferring latent topics. The proposed thLDA model is significant in that it can overcome the barrier that unsupervised models have when it is applied to noisy and sparse data. By transferring different domain knowledge, thLDA also saves time and  the cumbersome annotation efforts required for supervised models. thLDA has an advantage over LDA model by producing a topic hierarchy and document clusters without additionally computing similarity among topic distribution of documents. Furthermore, thLDA offers major advantages over other supervised or semisupervised LDA models by providing mixture of detailed a topic hierarchy below a certain level in an unsupervised way, while providing the above that level topic structure guided by given prior knowledge. By applying prior knowledge in both supervised and unsupervised ways, we can apply thLDA to learn deeper level of topic hierarchy than the given depth of source domain prior hierarchy. In the following experiment, we will show the performance of thLDA a combination of supervised prior knowledge upto a certain level and unsupervised below that level.

\section{Experiment Results}
In our experiments, we used one source domain and three target domain text data sets to show the effectiveness of our transfer hierarchical LDA model. We used two well known text collections: the Associated Press (AP) Data set\cite{harman1993first} and Reuters Corpus Volume 1(RCV1) Data set\cite{lewis2004rcv1} and one sparse and noisy Microblog data from Twitter for target domains and Yahoo! News categories for source domain. 

\begin{figure*}[tbh]
%\begin{left}
\begin{tabular}{p{7.5cm}p{7.5cm}}

\subfigure[hLDA]{
\includegraphics[width=1.05\linewidth]{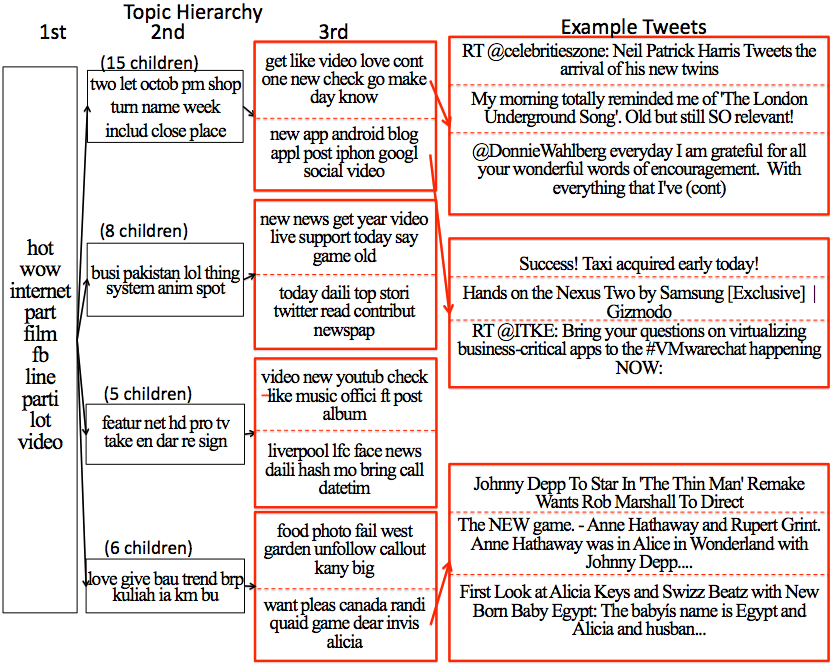} %hlda-tree.png}
\label{fig:hlda-tree}
}&
\subfigure[thLDA]{
\includegraphics[width=1.1\linewidth]{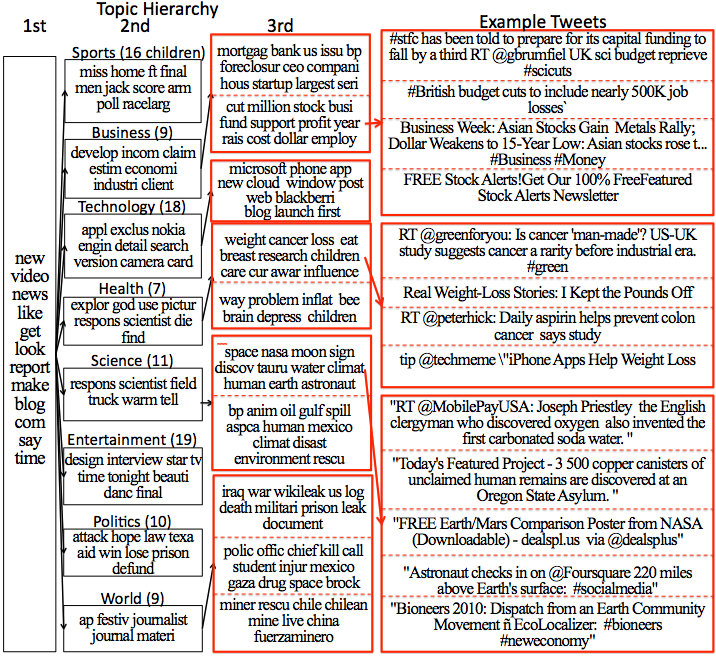} %thlda-tree.png} 
\label{fig:lhlda-tree} %
}
\\
\end{tabular}
%\end{left}
\caption{The topic hierarchy learned from (A) hLDA and (B) thLDA as well as example tweets assigned to the topic path.} 

\end{figure*}

\subsection{Dataset Description}
We used the web crawler to fetch news titles on 8 categories in Yahoo! News: science, business, health, sports, politics, world, technology and entertainment, as shown in Fig \ref{fig:lhlda} (a). We parsed and stemmed news titles and computed the tf-idf  score to generate weighted word vector for each topic category. We computed each topic category score using the top 50 tf-idf weighted word vector and picked one optimal category per level as a label for each target document.

Text Retrieval Conference AP (TREC-AP) \cite{harman1993first} contains Associated Press news stories from 1988 to 1990. The original data includes over 200,000 documents with 20 categories. The sample AP data set from \cite{blei2004hierarchical}, which is sampled from a subset of the TREC AP corpus contains D=  2,246 documents with a vocabulary size V = 10,473 unique terms. We divided documents into 1,796 observed documents  and 450 held-out documents to measure the held-out predictive log likelihood.

RCV1\cite{lewis2004rcv1} is an archive of over 800,000 manually categorized newswire stories provided by Reuters, Ltd. distributed as a set of on-line appendices to a JMLR article. It also includes 126 categories, associated industries and regions. 
For this work a subset of RCV1 data set is used. 
Sample RCV1 data set has D = 55,606 documents with a vocabulary size V = 8,625 unique terms. We randomly divided it into 44,486 observed documents and 11,120 held-out documents for experiments.  

We have crawled the Twitter data for two weeks and obtained around 58,000 user profiles and 2,000,000 tweets. Twitter users use some structure conventions such as a user-to-message relation (i.e. initial tweet authors, via, cc, by), type of message (i.e. Broadcast, conversation, or retweet messages), type of resources (i.e. URLs, hashtags, keywords) to overcome the 140 character limit. To capture trending topics, many applications analyze twitter data and group similar trending topics using structure information (i.e. hashtag or url) and shows a list of top N trending topics. However according to \cite{BoydGL10}, only 5\% of tweets contain a hashtag with 41\% of these containing a URL, and also only 22\% of tweets include a URL. In this work, instead of using structure information (i.e. hashtag or url) of a tweet, we used only words. We removed structural information such as initial tweet authors, via, cc, by, and url and stemmed word to transform any word into its base form using the Porter Stemmer.  For the experiment, we randomly sampled and used D = 21,364 documents with a vocabulary size V = 31,181 unique terms. We randomly divided it into 16,023 observed documents and 5,341 held-out documents.

\subsection{Performance Comparison on Topic Modeling}

\paragraph{LDA and Labeled LDA}

We implemented LDA and lLDA using the standard collapsed Gibbs Sampling method. To compare the learned topic results between supervised LDA and unsupervised LDA topic models, we ran standard LDA with 9, 20 and 50 topics and lLDA with 9 topics: Yahoo news 8 top level categories and 1 topic as freedom of topic. The results show two main observations: first, multiple topics from the standard LDA mapped to popular topics (i.e. technology, entertainment and sports) in lLDA; second, not all lLDA topics were discovered by standard LDA (i.e. topics such as politics, world, and science are not discovered).

 \paragraph{Hierarchical LDA}
We used HLDA-C \cite{blei2004hierarchical, blei2010nested}  with a fixed depth tree and a stick breaking prior on the depth weights. The topic hierarchy generated by hLDA is shown in Fig \ref{fig:hlda-tree}. Being an unsupervised model, the hLDA gives the result that totally depends on term co-occurance in the documents. hLDA gives a topic hierarchy that is not easily understood by human beings, because each tweet contains only small number of terms and co-occurance would be very sparse and less relevant compared to long documents. Nodes in the topic hierarchy capture some clusters of words from the input documents, such as the 2nd topic in 3rd column in Fig \ref{fig:hlda-tree}, which has key words focusing on smart phones (Android, iPhone, and Apple) and the 5th nodes in 3rd column that covers online multimedia resources. However, the 2nd level topic nodes are less informative and the relationship between child nodes in the 3rd level and their parent nodes in the 2nd level is less semantically meaningful. The topics belong to the same parent in level 3 usually do not relate to each other in our run result. Ideally, this should work for documents that are long in length and dense in word distribution and overlapping. However, hLDA gives less interpretable results on noisy and sparse data.

\begin{figure*}[tbh]
\begin{center}
\begin{tabular}{p{4.5cm}p{4.5cm}p{4.5cm}}

\includegraphics[width=1.0\linewidth]{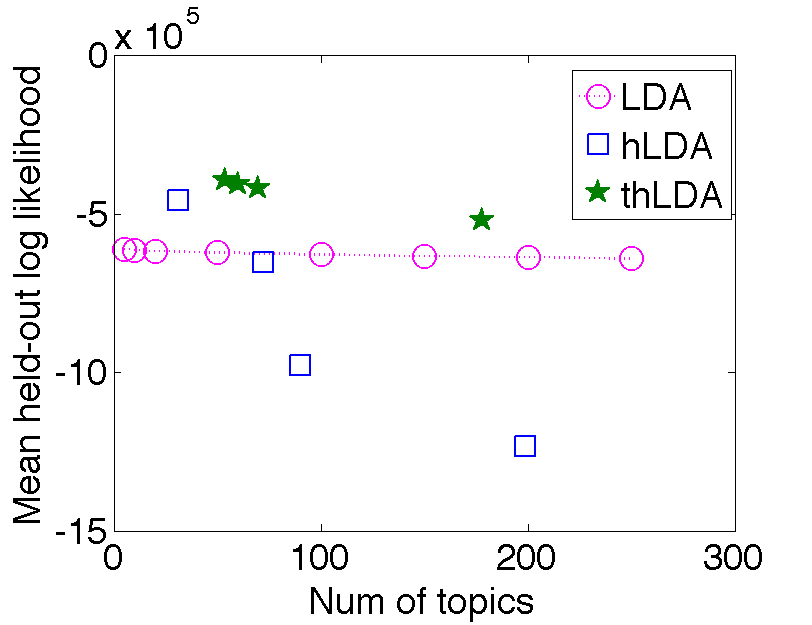}
&\includegraphics[width=1.0\linewidth]{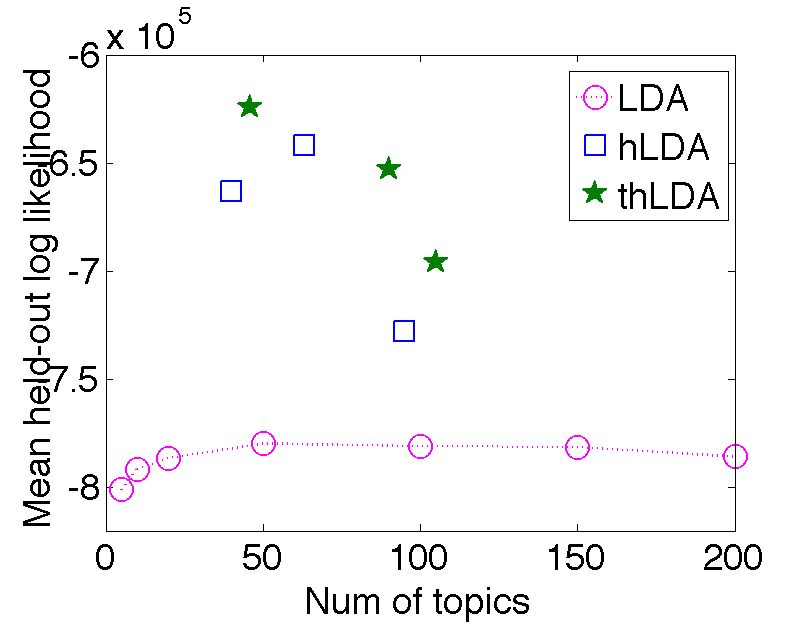} 
& \includegraphics[width=1.0\linewidth]{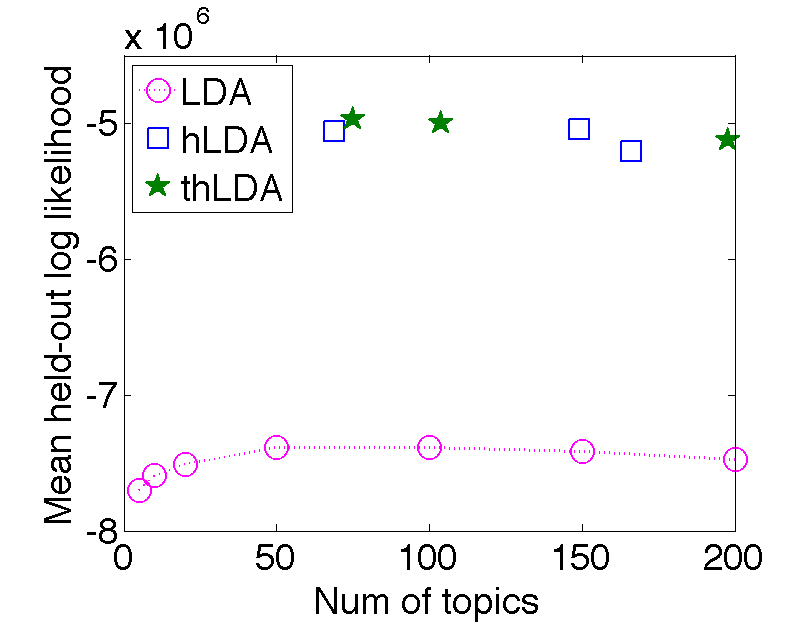}
\\

\centering(a) Twitter&\centering(b) AP&\centering(c) RCV1\\

\end{tabular}
\end{center}
\caption {The held-out predictive log likelihood comparison for LDA, hLDA, and thLDA model on three different data set}
\label{fig:heldOUT}
\end{figure*}

\paragraph{Transfer Hierarchical LDA}
We implemented standard thLDA by modifying HLDA-C. Fig \ref{fig:lhlda-tree} shows two important advantages of our model: first, topic nodes are better interpretable by transferring our source domain knowledge. Second, all the child topics reflect a strong semantic relationship with their parent. For example, in the topic "world", the first child "iraq war wikileak death" relates to the Iraq war topic, the second child "police kill injury drup" relates to criminals and more interestingly, the 3rd child topic "chile rescue miner" denotes the recent event of 33 traped miners in Chile. Note that we only impose prior knowledge on the 2nd level and all 3rd level topics are automatically emerged from the data set. 
In "science" topics the 1st child topic "space Nasa moon" is about astronomy and the 2nd child topic "BP oil gulf spill environment" is about the recent Gulf Oil Spill. Furthermore, example tweets assigned to the topic nodes show a strong association with tweet clustering.

To quantitatively measure performance of thLDA, we used predictive held-out likelihood. We divided the corpus into the observed and the held-out set, and approximate the conditional probability of the held-out set given the training set. To make a fair comparison we applied same hyper parameters that exist in all three models while applied a different hyper parameter $\eta$ to obtain a similar number of topics for hLDA and thLDA models. Following \cite{blei2010nested}, we used outer samples: taking samples 100 iterations apart using a burn-in of 200 samples. We collected 800 samples of the latent variables given the held-out documents and computed their conditional probability given the outer sample with the harmonic mean. Fig \ref{fig:heldOUT} illustrates the held-out likelihood for LDA, hLDA, and thLDA on Twitter, AP, and RCV1 corpus. In the figure, we applied a set of fixed topic cardinality on LDA and fixed depth 3 of the hierarchy on hLDA and thLDA. We see that thLDA always provides better predictive performance than LDA or hLDA on all three cases (Fig \ref{fig:heldOUT}). Interestingly, thLDA provides significantly better performance on (a) Twitter data set, while hLDA shows poor performance than LDA. As Blei et al \cite{blei2004hierarchical,blei2010nested} showed that eventually with large numbers of topics LDA will dominate hLDA and thLDA in predictive performance, however thLDA performs better in predictive performance with reasonable numbers of topics. 

For manual evaluation of tweet assignments on learned topics, we randomly selected 100 tweets and manually annotate correctness. For LDA, we pick the highest assigned topic and for hLDA, thLDA, and parallel-thLDA, third level topic node is used and their accuracy were 41\% 46\%, 71\%, and 56\% respectively. Table \ref{tbl:tweetex} shows example tweets with their assigned topic from LDA, hLDA, and thLDA.

\begin{figure*}[tbh]
\begin{center}
\begin{tabular}{cc}
\includegraphics[width=0.38\linewidth]{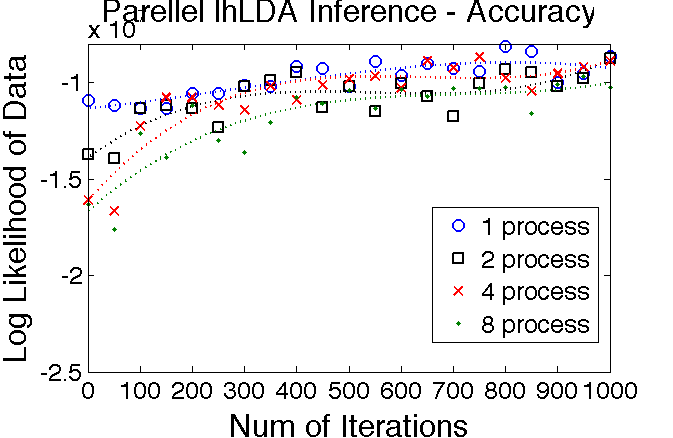}&
\includegraphics[width=0.38\linewidth]{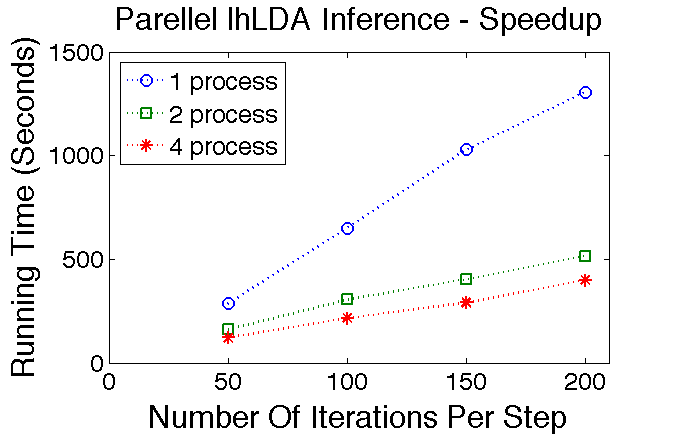}\\
(a)& (b)\\
\end{tabular}
\end{center}
\caption{(a)Parallel thLDA approximation inference performance comparison. (b)Parallel thLDA approximation inference speedup comparison.}
\label{fig:parallel_accuracy}
\end{figure*}

\subsection{Performance Comparison on Scalablity}

We evaluate our approximate parallel learning method on the twitter dataset with 5000 tweets. The log likelihood of training data during the Gibbs sampling iterations is shown in Fig \ref{fig:parallel_accuracy} (a). In all cases, the gibbs sampling converges to the distribution that has similar log likelyhood. Fig \ref{fig:parallel_accuracy} (b) shows the speedup from the parallel inference method. 
 In addition to the overhead of topic tree merging stage, the system also suffers from the overhead of state loading and saving time, which is similar since they occur everytime we need to update the global tree. Because of these overheads, the system performs better when merging step is large. When we merge topic trees in every 50 gibbs iterations, the speedup with 4 processes is 2.36 times faster than with 1 process, but when we merge topic trees in every 200 gibbs iterations, the speedup is 3.25 times. The overhead can be roughly seen if we extend the lines in Fig \ref{fig:parallel_accuracy} (b) to intersect with y-axes. Since the merging stage complexity is linear to the number of topic trees being merged, we see a greater overhead in 4 process experiment. However, the merging algorithm complexity does not depend on the size of dataset, which means the merging overhead will be ignorable when run with huge datasets.

\begin{table*}[]
\centering
\scriptsize{
\begin{tabular}{|p{5cm}|p{3cm}|p{3cm}|p{3cm}|}
\hline
\multicolumn{1}{|c|}{Tweet} & \multicolumn{1}{c|}{LDA} & \multicolumn{1}{c|}{hLDA} & \multicolumn{1}{c|}{thLDA} \\
\cline{1-4}
\multicolumn{1}{|c|}{\bf " \@usnoaagov: NOAA scientists monitor } & \multicolumn{1}{c|}{servic switch roll} & \multicolumn{1}{c|}{tcot teaparti palin} & \multicolumn{1}{c|}{climat discov check earth} \\
\multicolumn{1}{|c|}{\bf ozone levels above } & \multicolumn{1}{c|}{ broadband resid } & \multicolumn{1}{c|}{tea sgp parti gop} & \multicolumn{1}{c|}{ water chang sea found} \\
\multicolumn{1}{|c|}{\bf Antarctic: "} & \multicolumn{1}{c|}{voip ip chang asterisk} & \multicolumn{1}{c|}{democrat obama nvsen} & \multicolumn{1}{c|}{scientist arctic warm} \\
\cline{1-4}
\multicolumn{1}{|c|}{\bf FB RT: Breaking News: NATO official:} & \multicolumn{1}{c|}{call edit via limit } & \multicolumn{1}{c|}{one game yanke} & \multicolumn{1}{c|}{minist foreign pakistan} \\
\multicolumn{1}{|c|}{\bf Osama bin Laden is hiding} & \multicolumn{1}{c|}{duti topic live} & \multicolumn{1}{c|}{know new video} & \multicolumn{1}{c|}{israel un us gaza} \\
\multicolumn{1}{|c|}{\bf in northwest Pakistan. - } & \multicolumn{1}{c|}{playstat novel cafe} & \multicolumn{1}{c|}{find ranger make man} & \multicolumn{1}{c|}{secur aid call} \\
\cline{1-4}

\end{tabular}
}
\caption {Example Tweets and their assigned topic with top 10 words from LDA, hLDA, and thLDA model.}
\label{tbl:tweetex}
\end{table*}

\section{Conclusion}
In this paper, we proposed a transfer learning approach for effective and efficient topic modeling analysis on social media data. More specifically, we developed transfer hierarchical LDA model, an extension of hierarchical LDA model, which inferred the topic distributions of documents while incorporating knowledge from other domains. In addition, we designed a parallel inference framework to run parallel Gibbs sampler synchronously on multi-core machines so as to perform topic modeling on large-scale datasets. Our work is significant in that it is among the frontier approaches to explore knowledge transfer from other domains to topic modeling for large-scale microblog analysis. For future work, we are interested in exploring other effective approaches to transfer domain knowledge in addition to topic priors as examined in the current paper.

\section*{Acknowledgment}

Research was sponsored by the U.S. Defense Advanced Research Projects
Agency (DARPA) under the Anomaly Detection at Multiple Scales (ADAMS)
program, Agreement Number W911NF-11-C-0200. The views and conclusions
contained in this document are those of the author(s) and should not
be interpreted as representing the official policies, either expressed
or implied, of the U.S. Defense Advanced Research Projects Agency or
the U.S. Government. The U.S. Government is authorized to reproduce
and distribute reprints for Government purposes notwithstanding any
copyright notation hereon.

\bibliographystyle{abbrv}
\bibliography{reference}
\end{document}